\documentclass{article}

\PassOptionsToPackage{numbers, compress}{natbib}

\usepackage[final]{nips_2018}

\usepackage[utf8]{inputenc} 
\usepackage[T1]{fontenc}    
\usepackage[hidelinks]{hyperref}       
\usepackage{url}            
\usepackage{booktabs}       
\usepackage{amsfonts}       
\usepackage{nicefrac}       
\usepackage{microtype}      
\usepackage[acronym]{glossaries} 
\usepackage{graphicx}

\title{Binary Input Layer: Training of CNN models with binary input data}

\author{
  Robert D\"urichen$^1$, Thomas Rocznik$^2$, Oliver Renz$^1$, Christian Peters$^2$
 \\
  $^1$ Robert Bosch GmbH, Renningen, Germany\\
  $^2$ Robert Bosch LLC, Research and Technology Center, Sunnyvale, CA, USA\\
  \texttt{robert.duerichen@de.bosch.com} \\
}

\newacronym{cnn}{CNN}{convolutional neural networks}
\newacronym{loso}{LOSO}{leave one subject out}
\newacronym{pamap2}{PAMAP2}{physical activity monitoring}
\newacronym{svhn}{SVHN}{street view house number}
\newacronym{cifar}{CIFAR-10}{CIFAR-10}
\newacronym{bil}{BIL}{binary input layer}
\newacronym{dbi}{DBID}{direct binary input data}
\newacronym{pp}{pp}{percentage points}
\newacronym{fc}{FC}{fully connected}
\newacronym{fpid}{FPID}{fixed point input data}

\usepackage{xcolor}

\begin{document}

\maketitle

\begin{abstract}
For the efficient execution of deep \gls{cnn} on edge devices, various approaches have been presented which reduce the bit width of the network parameters down to $1\,$bit.
Binarization of the first layer was always excluded, as it leads to a significant error increase.
Here, we present the novel concept of \gls{bil}, which allows the usage of binary input data by learning bit specific binary weights. 
The concept is evaluated on three datasets (\acrshort{pamap2}, \acrshort{svhn}, \acrshort{cifar}).
Our results show that this approach is in particular beneficial for multimodal datasets (\acrshort{pamap2})	 where
it outperforms networks using full precision weights in the first layer by $1.92\,$\gls{pp} while consuming only $2\,\%$ of the chip area.
\end{abstract}

\section{Introduction}
Deep \gls{cnn} have become a widely used standard machine learning method for a wide variety of domains such as computer vision and speech recognition. 
However, these models often require large amount of memory to store all parameters and have a high computational complexity.
Different approaches have been investigated by either designing novel network architectures  \cite{Howard2017} or by simplifying existing networks using, for instance, low bit width quantization of network parameters \cite{Li2016,Zhou2016,Rastegari2016} to enable inference on wearables or other edge devices.\\
Courbariaux \textit{et al.} \cite{Courbariaux2016a} showed that by using binarized weights and activations, almost state-of-the-art performance can be achieved on the \acrlong{cifar} and \gls{svhn} dataset.  
On the more challenging ImageNet dataset, Rastegari \textit{et al.} \cite{Rastegari2016} presented a binary weight network which achieved an accuracy comparable to a full-precision network. 
By binarizing additionally the activations, the Top-1 accuracy drops by 12.4 \gls{pp} to $44.2\,\%$.
DOREFA-NET by Zhou \textit{et al.} \cite{Zhou2016} extended this idea by additionally using low bit width gradients, which is in particular relevant for online training on edge devices.\\
All approaches have in common that they exclude binarization of the first (and last) layer. 
In general, the number of parameters required and computations executed within the first layer are relatively low compared to the residual network. 
The input data have typically much fewer channels (e.g. for color images only: red, green, and blue) than representations within the residual network (e.g. \gls{cnn} layer with 512 filters). 
The consequence is that two different type of multipliers are needed with different bit widths to perform all computations.
However, for some applications a completely binarized system is beneficial to further minimize energy consumption and cost (e.g. wearables such as sensor patches). 
First results toward such a solution were presented in \cite{Zhou2016}. 
The authors used 1, 2, 4 bit resolutions for weights, activations and gradients, respectively, for the complete network (incl. first layer) and reported an accuracy decrease of  $0.3\,\gls{pp}$ evaluated on \acrlong{cifar}.
Nonetheless, the computations performed at the first layer cannot be executed by a binary multiplier as non-binary input data with $M$ bit precision was multiplied by binary weights. \\
We envision a system where the complete network can be executed in binary format. 
Here, the concept of a \gls{bil} is presented which allows efficient binary computation of all multiplications at the input layer.
This paper makes the following contributions:
\begin{enumerate}
	\item We present a general approach to realize binary computations within the first layer of a \gls{cnn} model by using \gls{bil}. We use bit specific binary weigths which enables the models to learn which bits of the input data are relevant and which not.
	\item We evaluate our concept on two image classification datasets (\gls{svhn} and \acrlong{cifar}) and on a wearable based human activity recognition dataset (\acrshort{pamap2}). Latter is in particular interesting as it contains multimodal input data such as inertial data which are often used in very low power edge devices like wearables.	
\end{enumerate}

\section{Methods}
\begin{figure}
	\centering
	\includegraphics[width=\linewidth]{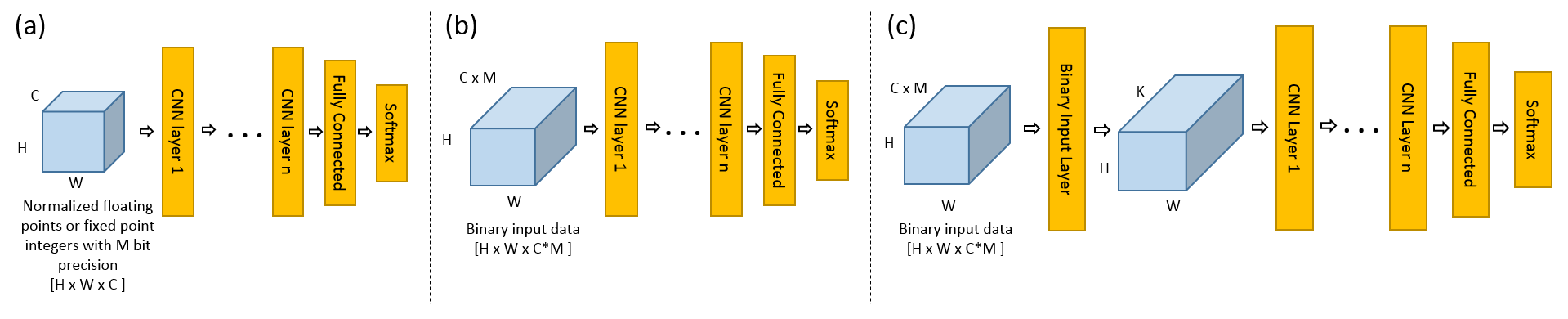}
	\caption{Illustration of a typical \gls{cnn} model consisting of $n$ \gls{cnn} layer followed by a fully connected and softmax layer using different kind of input data: (a) fixed point integer values with $M$ bit precision, (b) binary input data directly (\gls{dbi}), and (c) binary input data and an additional binary input layer \gls{bil} with $K$ filters. The original input data has the dimension of $H\times W \times C$.}
	\label{fig:networks}
\end{figure}
A typical \gls{cnn} model is illustrated in Figure \ref{fig:networks}(a) with $n$ convolutional layers followed by a \gls{fc} and softmax layer.  
We assume that the input data has a shape of $H \times W \times C$, with $C$ being the number of channels (i.e. 3 channels in case of color images).
Each input value $x$ is a fixed-point integer with $M$ bit precision s.t. $x = \sum_{m=0}^{M-1}x^b_m \cdot 2^{m} $, with $x^b_m$ being the $m$th bit of $x$. 
Typical bit resolutions are in the range from $M=8\,$bit for imaging problems but up to $16\,$bit for inertial and environmental sensor data. 
In a standard processing chain, the input data are often normalized to e.g. [0,1].
This results in a conversion of the input data from $M$ bit fixed-point integers to floats.
In binarized \glspl{cnn}, these floating point inputs are multiplied by floating point weights before some clipping is applied to a binarize the values \cite{Zhou2016,Courbariaux2016a,Rastegari2016}. 
We refer to this approach as baseline. \\
A straight forward approach to binarize multiplications at the first layer was discussed in \cite{Zhou2016}: 
\begin{equation}
s = \sum_{m=0}^{M-1} 2^{m}(x^b_m \cdot w^b),
\end{equation}
with $w^b$ being a binary weight and $s$ being the resulting sum. 
Here, the normalization step is removed and the multiplication is performed using $M$ bit fixed-point integers. 
The limitation of this approach is that all $x^b_m$ are multiplied by the same binary weight $w^b$. 
Further, the ordinal structure of the input bits is defined by the factor $2^m$ which also results in additional computations. 
We refer to this approach as \glsreset{fpid}\gls{fpid}.\\
We extend this approach by introducing bit specific weights $w^b_m$ such that 
\begin{equation}
s = \sum_{m=0}^{M-1} (x^b_m \cdot w^b_m).
\label{bitspecific}
\end{equation}
The advantage of this approach is that the algorithm can learn a bit specific relevance without a predefined ordinal structure.
As all input data are binary, no further normalization of the input data is required. 
We refer to this approach as \glsreset{dbi}\gls{dbi}. \\
We extend this idea further by introducing a \acrlong{bil} (\acrshort{bil}), which is an additional convolutional filter layer between the binary input data and the first layer of the \gls{cnn} model with $K$ filters and a filter size of $1 \times 1$.  
This approach is motivated by MobilNets \cite{Howard2017}, which factorizes a standard convolution in a depthwise and pointwise convolution. 
Our appraoch is illustrated in Figure \ref{fig:networks}(c). 
\begin{table}[t]
	\caption{Comparision of the first layer computations for the four approaches: Baseline, \gls{fpid}, \gls{dbi}, and \gls{bil} with respect to variable type and  number of multiplications and weights. We assume that the input data has a shape $H \times W \times C$ and  a precision of $M$ bits, the first convolutional layer has $I$ filters with size $F \times F$, and that \gls{bil} has $K$ filter. Relative area and energy consumption compared to baseline estimated for the \gls{pamap2} model ($K=64$).}
	\label{comp-table}
	\centering
	\begin{tabular}{lllll}
		\toprule
		& Baseline    		&  \gls{fpid} & \gls{dbi} & \gls{bil}\\
		\midrule
		Input type 					& floating points 	& fixed point int.    	& binary 	& binary    \\
		Weight type 				& floating points 	& binary  				  	& binary 	& binary    \\
		\# multiplications   		& $H W C F^2 I$					& $H W C F^2 I$ & $H W C F^2 I  M$ & $H\,W\,K (M\, C  + F^2 I)$       \\
		\# weights					& $C F^2 I$	& $C F^2 I$			& $C F^2 I M$	& $C M K + F^2 I K$ \\
		\midrule
		Rel. area \& energie			&	$100\,\%$		& $0.21\,\%$& $0.21\,\%$	& $1.86\,\%$	\\
		\bottomrule
	\end{tabular}
\end{table}
Table \ref{comp-table} lists a comparison between the four approaches discussed. 

\section{Datasets and Models}
We evaluate our concept using the DoReFa-net framework \cite{Zhou2016} and the tensorpack implementation\footnote{https://github.com/tensorpack/tensorpack}. 
Binarization of the weights is based on the “straight-through estimator” method using a constant scalar to scale all filters ($w^b = sign(w^i) \times E(|w^i|)$ with $E(|w^i|)$ being the mean of all absolute values). 
After the computed sum $s$ has passed through a bounded activation function $h$ which ensures $a^i \in [0,1]$, the output activation is quantized by $a^b = quantize_k (a^i)$ which converts $a^i$ into a $k$ bit integer (here $k=1$). 
The bit width of the gradients remains for all experiments $32$, as our focus lies on inference. 
For all approaches the training was stopped after $200$ epochs and the accuracies are reported.
All models were trained using ADAM optimization with an initial learning rate of $10^{-4}$ for \gls{pamap2} and $10^{-3}$ for \gls{svhn} and \acrlong{cifar} using a decreasing step function. 
The \acrlong{bil} is realized as \textit{convolution - batch norm - nonl. activation}. All input data was converted to $M=8\,$bit for \gls{fpid}, \gls{dbi}, and \gls{bil}.

\paragraph{PAMAP2}
The \gls{pamap2} dataset is a human activity recognition dataset based on multimodal wearable sensors containing 18 activities \cite{Reiss2012}. 
We follow the implementation of \cite{Muenzner2017} and arranged all activities in 7 groups (lying, sitting, standing, walking, stair climbing, sport, house work). 
A smart watch like behavior is emulated by using only data of the wrist sensor (3D accelerometer and gyroscope) and heart rate monitor (7 dimensions in total). 
The inputs are $1\,s$ long time windows (resulting in $100$ values with a sampling rate of $f_s =100\,Hz$).
The input dimension is 7$\times$100$\times$1.
We use the network of \cite{Muenzner2017} and extend it by additional max pooling (MP) layers to: \textit{24-C3+MP2+32-C3+MP2+64-C64+MP2+FC256+Softmax}.
Note for this dataset, convolutions and max pooling are only performed along the time dimension (C3 is a filter of shape $[1\times3]$, MP2 has a shape $[1\times2]$).
Batch normalization is added after each layer and a drop out layer before the FC layer. 
The evaluation is done using a \gls{loso} cross evaluation of the first 8 subjects.
\paragraph{SVHN} 
The \gls{svhn} dataset contains photos from house numbers. 
Our evaluation is based on the "cropped" format of the dataset with 32$\times$32 pixel colored images centered around each digit. 
Both subsets ("training" and "extra") of \gls{svhn} dataset were used as training data. 
We followed the implementation of \cite{Zhou2016} using model A (7 layer network): \textit{48-C5+MP2-2$\times$(64-C3)-MP2-3$\times$(128-C3)-FC512-Softmax} with batch normalization at each layer and a drop out layer before the FC layer. 
All pictures were resized to $40\times40$ pixel.
\paragraph{CIFAR-10}
We follow the VGG inspired architecture of \cite{Li2016} for \acrlong{cifar}, which consists of \textit{$2\times(128-$C3$)$+MP2 + $2\times(256-$C3$)$ +  MP2 + $2\times(512-$C3$)$ + MP2 + $1024-$FC + Softmax}.
The size of the training data was increased using data augmentation similar to \cite{Li2016} (resizing and randomly cropping of the image and horizontal flipping). 
\begin{table}[t]
	\caption{Validation error in [$\%$] of networks with only full precision input data and weights at the first layer (baseline), \gls{fpid}, \gls{dbi}, and \gls{bil} approach with different number of filters $K$ for \gls{pamap2}, \gls{svhn}, and \gls{cifar}. Only the last layer of each model is not quantized to 1bit.}
	\label{error_table}
	\centering
	\begin{tabular}{lccccccc}
		\toprule         
		Dataset   			& Baseline     	& \gls{fpid} 		& \gls{dbi} & \gls{bil} & \gls{bil} & \gls{bil} & \gls{bil} \\
							&				&				&			& (K=32)	& (K=64)	& (K=128) 	&(K=256)\\
		\midrule
		\gls{pamap2}		& 21.66			& 40.62			&  25.48	& 27.51 	& 21.45		& 19.73		& 19.74	\\
		\gls{svhn}			& 3.15 (2.53\cite{Courbariaux2016a})		& 4.33		&  3.8		& 4.32 		& 4.11		& 3.61		& 3.42	\\
		\acrlong{cifar}		& 9.07 (9.71\cite{Zhou2016})	& 10.94			&  15.48	&  35.27	&  18.29	&  16.37	&  13.65\\
		\bottomrule
	\end{tabular}
\end{table}

\section{Results and Discussion}
Table \ref{error_table} shows the validation error for the discussed approaches on all three datasets. 
In principle, similar tendencies can be observed for \gls{pamap2} and \gls{svhn}. 
Using the \gls{fpid} or \gls{dbi} approach increases the validation error significantly for \gls{pamap2} and slightly for \gls{svhn}. 
By adding a \gls{bil} layer with $K\ge64$, the error can be decreased to the level of the baseline model besides \acrlong{cifar}. 
In case of the \gls{pamap2}, \gls{bil} can even outperform the baseline error by $1.92\,$\gls{pp}. 
The reasons for the good performance are two folded in our opinion.
First, \gls{pamap2} is a multimodal dataset containing accelerometer, gyroscope and heart rate data.
In contrast to image data, not the complete range of all sensor modalities is used (e.g. very high accelerations).
Consequently, not all input bits are equally relevant which motivates the use of bit specific binary weights. 
Second, the channels are treated differently for \gls{pamap2} and the image datasets. 
In case of \gls{pamap2}, the input data has the shape $[7\times100\times1]$ while for \gls{svhn} it has the shape $[40\times40\times3]$ and $[32\times32\times3]$ for \acrlong{cifar}. 
The consequence is that all (RGB) channels will be fused at the first layer in case of \gls{svhn} and \acrlong{cifar}, while they remain separately for \gls{pamap2}.
As the number of \gls{bil} filters $K$ is the same for all datasets, the image classification models have basically less parameters to learn a meaningful representation of the input data. 
This becomes clearly visible in case of \acrlong{cifar}, where the validation error of \gls{bil} with $K=256$ is $4.58\,\gls{pp}$ higher compared to the baseline model. 
We assume that the validation error of \gls{svhn} and \acrlong{cifar} can be further decreased by increasing the number of \gls{bil} filters or implementation of a channel independent network.
Another reason of the high error in case of \acrshort{cifar} is the small size of the data set, which is $12\times$ smaller as \gls{svhn} (without data augmentation).
Evaluating the training curves reveals that in particular for \gls{bil} the validation error shows high fluctuations compared to the training error. \\
Finally, we want to provide a fist impression on the hardware consequences for our approaches, as the main motivation for the binarization of the first layer is to reduce area and energy consumption.
We estimated the occupied ASIC area for each approach using the \gls{pamap2} model (with $K=64$) using  
a NAND gate estimation for all multiplications. 
The results have been normalized to the baseline without reuse of components or resource sharing.
Thus, a similar reduction of the energy consumption can be expected.  
Table \ref{comp-table}, shows that the relative space and energy for the BIL implementation decreases by $98.14\,$\gls{pp} with an error reduction of $1.92\,$\gls{pp} over baseline implementation at the same time.
\medskip

\small
\bibliographystyle{plainnat}
\bibliography{references}

\end{document}